\documentclass[10pt,twocolumn,letterpaper]{article}

\usepackage{cvpr}              

\usepackage[dvipsnames]{xcolor}

\usepackage[accsupp]{axessibility} 

\usepackage{graphicx}
\usepackage{amsmath}
\usepackage{amssymb}
\usepackage{booktabs}

\usepackage{bm}
\usepackage{colortbl}
\usepackage{times}
\usepackage{epsfig}
\usepackage{amssymb}
\usepackage{multirow}
\usepackage{threeparttable}
\usepackage{caption}
\usepackage{soul}
\usepackage{xcolor}
\usepackage{multirow} 

\definecolor{cvprblue}{rgb}{0.21,0.49,0.74}
\usepackage[pagebackref,breaklinks,colorlinks,citecolor=cvprblue]{hyperref}

\def\paperID{15454} 
\def\confName{CVPR}
\def\confYear{2024}

\title{DriveWorld: 4D Pre-trained Scene Understanding via World Models for Autonomous Driving}

\author{Chen Min$^{1,2}$, Dawei Zhao$^{2}$\footnotemark[1] , Liang Xiao$^{2}$\footnotemark[1] , Jian Zhao$^{3}$, Xinli Xu$^{4}$, Zheng Zhu$^{5}$\\ Lei Jin$^{6}$, Jianshu Li$^{7}$, Yulan Guo$^{8}$, Junliang Xing$^{9}$, Liping Jing$^{10}$, Yiming Nie$^{2}$, Bin Dai$^{2}$\\
$^{1}$School of Computer Science, Peking University \\ $^{2}$Unmanned Systems Technology Research Center, Defense Innovation Institute \\$^{3}$China Telecom Institute of AI \& NPU $^{4}$HKUST $^{5}$GigaAI $^{6}$BUPT $^{7}$Ant Group $^{8}$SYSU $^{9}$THU $^{10}$BJTU\\
{\tt\small minchen@stu.pku.edu.cn, adamzdw@163.com, xiaoliang@nudt.edu.cn}
}

\begin{document}
\maketitle

\renewcommand{\thefootnote}{\fnsymbol{footnote}} 
\footnotetext[1]{Corresponding authors.}

\begin{abstract}
Vision-centric autonomous driving has recently raised wide attention due to its lower cost. Pre-training is essential for extracting a universal representation. However, current vision-centric pre-training typically relies on either 2D or 3D pre-text tasks, overlooking the temporal characteristics of autonomous driving as a 4D scene understanding task. In this paper, we address this challenge by introducing a world model-based autonomous driving 4D representation learning framework, dubbed \emph{DriveWorld}, which is capable of pre-training from multi-camera driving videos in a spatio-temporal fashion. Specifically, we propose a Memory State-Space Model for spatio-temporal modelling, which consists of a Dynamic Memory Bank module for learning temporal-aware latent dynamics to predict future changes and a Static Scene Propagation module for learning spatial-aware latent statics to offer comprehensive scene contexts. We additionally introduce a Task Prompt to decouple task-aware features for various downstream tasks. The experiments demonstrate that DriveWorld delivers promising results on various autonomous driving tasks. When pre-trained with the OpenScene dataset, DriveWorld achieves a 7.5\% increase in mAP for 3D object detection, a 3.0\% increase in IoU for online mapping, a 5.0\% increase in AMOTA for multi-object tracking, a 0.1m decrease in minADE for motion forecasting, a 3.0\% increase in IoU for occupancy prediction, and a 0.34m reduction in average L2 error for planning. 
\end{abstract}    
\section{Introduction}
\label{sec:intro}

\begin{figure}[t]
	\centering
	\includegraphics[width=0.4\textwidth]{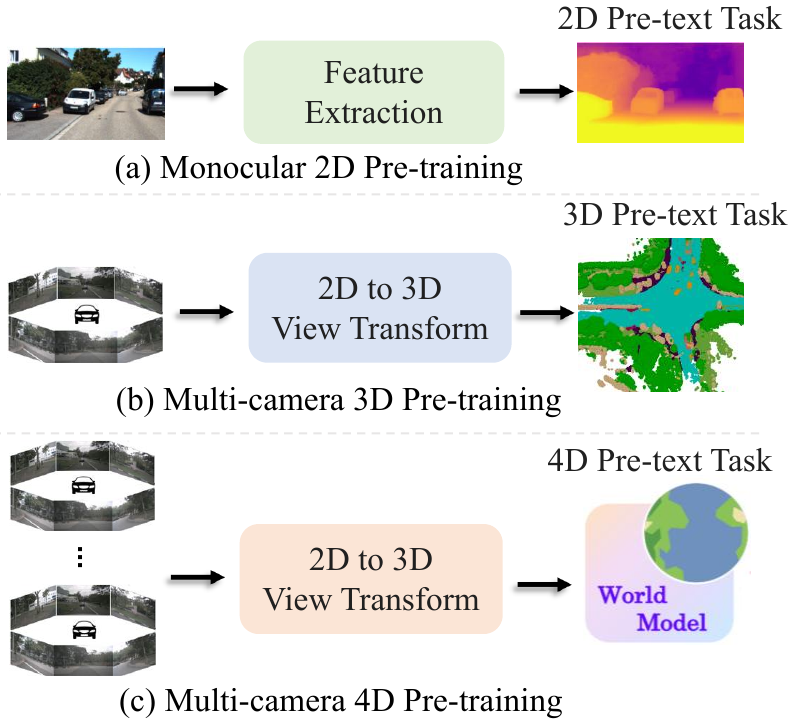} 
	\caption{Comparison on different pre-training methods for vision-centric autonomous driving. (a) Monocular 2D pre-training with 2D pre-text tasks (\eg, 2D classification and depth estimation). (b) Multi-camera 3D pre-training via 3D scene reconstruction or 3D object detection. (c) The proposed 4D pre-training based on world models learns unified spatio-temporal representations.}
	\label{fig:compare1}
\end{figure}

Autonomous driving is a complex undertaking that relies on comprehensive 4D scene understanding~\cite{implicit,driveanywhere}. This demands the acquisition of a robust spatio-temporal representation that can address tasks involving perception, prediction, and planning~\cite{uniad}. Learning spatio-temporal representations is highly challenging due to the stochastic nature of natural scenes, the partial observability of the environment, and the diversity of downstream tasks~\cite{wu2023spatiotemporal,nunes2023temporal}. Pre-training plays a crucial role in acquiring a universal representation from massive data, enabling the construction of a foundational model enriched with common knowledge~\cite{bert,rebuffi2017learning,ponderv2,pimae,wang2023mvcontrast}.
However, research on pre-training for spatio-temporal representation learning in autonomous driving remains relatively limited.

Vision-centric autonomous driving has recently attracted increasing attention because of its lower cost~\cite{bevformer,bevdet,uniad,detr3d,petr,bevdepth,beverse}. However, as shown in Fig.~\ref{fig:compare1}, the existing vision-centric pre-training algorithms still predominantly rely on 2D pre-text tasks~\cite{resnet,dd3d} or 3D pre-text tasks~\cite{occnet,uniscene,unipad}. DD3D~\cite{dd3d} has demonstrated the effectiveness of depth estimation for pre-training. OccNet~\cite{occnet}, UniScene~\cite{uniscene}, and UniPAD~\cite{unipad} have further extended pre-training to 3D scene reconstruction. However, these algorithms overlook the importance of 4D representation for understanding self-driving scenes. 

We aim to employ world models to address 4D  representation for vision-centric autonomous driving pre-training. World models excel in representing an agent's spatio-temporal knowledge about its environment.~\cite{lecun,world_models}. In reinforcement learning, DreamerV1~\cite{dreamerv1}, DreamerV2~\cite{dreamerv2}, and DreamerV3~\cite{dreamerv3} employ world models to encapsulate an agent's experience within a predictive model, thereby facilitating the acquisition of a wide array of behaviours. MILE~\cite{mile} leverages 3D geometry as an inductive bias and learns a compact latent space directly from videos of expert demonstrations to construct world models in the CARLA simulator~\cite{carla}. ContextWM~\cite{contextwm} and SWIM~\cite{swim} pre-train world models with abundant in-the-wild videos to enhance the efficient learning of downstream visual tasks. More recently, GAIA-1~\cite{gaia} and DriveDreamer~\cite{drivedreamer} have constructed generative world models that harness video, text, and action inputs to create realistic driving scenarios using diffusion models. Unlike the aforementioned prior works on world models, our approach primarily focuses on harnessing world models to learn 4D representations for autonomous driving pre-training.

Driving inherently entails grappling with uncertainty~\cite{hu}. There are two types of uncertainty in ambiguous autonomous driving scenarios: aleatoric uncertainty, stemming from the stochastic nature of the world, and epistemic uncertainty, arising from imperfect knowledge or information~\cite{uncertainty}. How to leverage past experience to predict plausible future states, and estimate missing information about the state of the world for autonomous driving remains an open problem. In this work, we explore 4D pre-training via world models to deal with both aleatoric and epistemic uncertainties. Specifically, we design the Memory State-Space Model to reduce uncertainty within autonomous driving from two aspects. Firstly, to address aleatoric uncertainty, we propose the Dynamic Memory Bank module for learning temporal-aware latent dynamics to predict future states. Secondly, to mitigate epistemic uncertainty, we propose the Static Scene Propagation module for learning spatial-aware latent statics to provide comprehensive scene context. Furthermore, we introduce Task Prompt, which leverages semantic cues as prompts to tune the feature extraction network adaptively for different driving downstream tasks. 

To validate the performance of our proposed 4D pre-training approach, we conducted pre-training on the nuScenes~\cite{nuscenes} training set and the recently released large-scale 3D occupancy datasets, OpenScene~\cite{openscene}, followed by fine-tuning on the nuScenes training set. The experimental results demonstrate the superiority of our 4D pre-training approach when compared to 2D ImageNet pre-training~\cite{resnet}, 3D occupancy pre-training~\cite{occnet,uniscene}, and knowledge distillation algorithms~\cite{bevdistill}. Our 4D pre-training algorithm exhibited substantial improvements in vision-centric autonomous driving tasks, including 3D object detection, multi-object tracking, online mapping, motion forecasting, occupancy prediction, and planning. The main contributions of this work are listed below:
\begin{itemize}
	\item We present the first 4D pre-training method based on world models for real-world vision-centric autonomous driving, which learns a compact spatio-temporal representation from multi-camera driving videos.
	\item We design the Memory State-Space Model, which includes a Dynamic Memory Bank module for learning temporal-aware latent dynamics, a Static Scene Propagation module for learning spatial-aware latent statics, and a Task Prompt to condition feature extraction adaptively for various tasks.
	\item Extensive experiments indicate DriveWorld's pre-training aids in establishing new state-of-the-art performance in vision-centric perception, prediction, and planning tasks.
\end{itemize}

\begin{figure*}[t]
	\centering
	\includegraphics[width=0.86\textwidth]{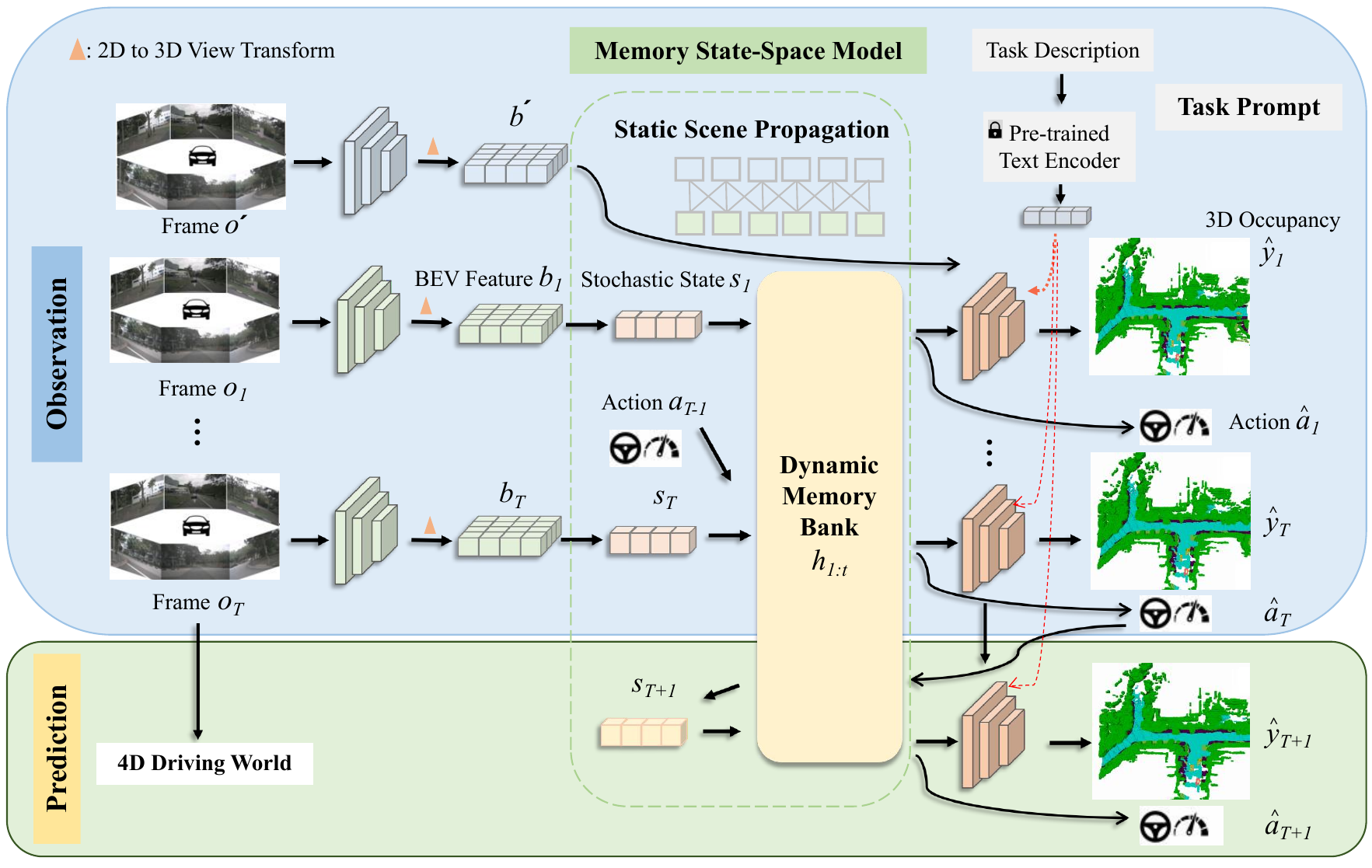}
	\caption{\small{Overall framework of the proposed DriveWorld. Since autonomous driving heavily relies on the understanding of 4D scenes, our approach first involves the transformation of multi-camera images into a 4D space. Within the proposed Memory State-Space Model for spatio-temporal modelling, we have two essential components: the Dynamic Memory Bank, which learns temporal-aware latent dynamics for predicting future states, and the Static Scene Propagation, which learns spatial-aware latent statics to provide comprehensive scene context. This configuration facilitates the decoder's task of reconstructing 3D occupancy and actions for both the current and future time steps. Besides, we design the Task Prompt based on a pre-trained text encoder to adaptively decouple task-aware features for various tasks.}
	}
	\label{fig:flowchart}
\end{figure*}

\section{Related Work}
\subsection{Pre-training for Autonomous Driving}
Based on input modalities, autonomous driving pre-training algorithms can be primarily categorized: pre-training on large-scale LiDAR point clouds~\cite{ad-pt,fei2023self,uni3d} and pre-training on images~\cite{liang2022effective,survey2,chen2021multisiam}. Pre-training algorithms for large-scale LiDAR point clouds can further be classified into contrastive learning methods~\cite{strl,gcc3d,proposalcontrast,bevcontrast,sheng2023contrastive,chen2022co}, masked autoencoder methods~\cite{occupancy-mae,voxel-mae,gd-mae,geomae,mv-jar}, and occupancy-based approaches~\cite{iscc,occupancy-mae,also,spot}. To incorporate 3D spatial structure into vision-centric autonomous driving, pre-training methods involving depth estimation have seen widespread adoption~\cite{dd3d,ppgeo}. OccNet~\cite{occnet}, UniScene~\cite{uniscene}, UniPAD~\cite{unipad}, and PonderV2~\cite{ponderv2} have introduced pre-training via 3D scene reconstruction. BEVDistill~\cite{bevdistill}, DistillBEV~\cite{distillbev}, and GeoMIM~\cite{geomim} employ knowledge distillation to transfer geometric insights from pre-trained LiDAR point clouds detection models. However, autonomous driving presents a 4D scene understanding challenge. We propose the first 4D pre-training approach based on world models for vision-centric autonomous driving.
\subsection{Spatial-Temporal Modeling for Autonomous Driving}
In the domain of autonomous driving, there has been significant research focus on spatio-temporal modeling. BEVFormer~\cite{bevformer} employs spatio-temporal transformers to learn BEV representations from multi-camera images. BEVDet4D~\cite{bevdet4d} extends BEVDet~\cite{bevdet} from spatial-only 3D space to the spatio-temporal 4D space. BEVStereo~\cite{bevstereo}, STS~\cite{sts}, and SOLOFusion~\cite{solofusion} address depth perception challenges in camera-based 3D tasks by leveraging temporal multi-view stereo (MVS)~\cite{mvs}. PETRv2~\cite{petrv2} and StreamPETR~\cite{streampetr} utilize sparse object queries to model moving objects and enable efficient transmission of long-term temporal information. ST-P3~\cite{stp3} and UniAD~\cite{uniad} are dedicated to building end-to-end vision-based autonomous driving systems through spatio-temporal feature learning.

\subsection{World Models}
World models enable intelligent agents to learn a state representation from past experiences and current observations, allowing them to predict future outcomes~\cite{world_models,lecun,dreamerv2}.
World models find extensive applications in reinforcement learning~\cite{dreamerv2,seo2023multi,pan2022iso}, and autonomous driving~\cite{gaia,wavbi,bogdoll2023exploring,gao2022enhance}. In reinforcement learning, \citet{world_models} proposed that the world model can be trained quickly in an unsupervised manner to learn a compressed spatial and temporal representation of the environment. Methods in ~\cite{recurrent_wm,dreamerv1,dreamerv2,dreamerv3} presuppose access to rewards and online interaction with the environment from predictions in the
compact latent space of a world model. ContextWM~\cite{contextwm} and SWIM~\cite{swim} pre-train world models with abundant in-the-wild videos for downstream visual tasks. 
In autonomous driving, \citet{occupancy} proposed the geometric occupancy grid as a world model for robot perception and navigation in 1989. MILE~\cite{mile} proposed to build the world model by predicting the future BEV segmentation from high-resolution videos of expert demonstrations for autonomous driving. GAIA-1~\cite{gaia} and DriveDreamer~\cite{drivedreamer} have constructed generative world models that harness video, text, and action inputs to create realistic driving scenarios using diffusion models. \citet{wavbi} builds unsupervised world models for the point cloud forecasting task in autonomous driving.
In this paper, we imbue the robot with a pre-trained spatio-temporal representation via world models to perceive surroundings and predict the future behaviour of other participants.
\section{DriveWorld}	
Consider a sequence of observed $T$ video frames denoted as $o_{1:T}$, captured by multi-view cameras, along with their corresponding expert actions, $a_{1:T}$ and 3D occupancy labels $y_{1:T}$ which can be acquired with the aid of LiDAR point clouds and pose data, we aim to learn a compact spatio-temporal BEV representation via world model that predicts current and future 3D occupancy given the past multi-camera images and actions. As shown in Fig.~\ref{fig:flowchart}, the designed world model consists of an Image Encoder, a 2D to 3D View Transform (\eg, Transformers~\cite{detr3d}, LSS~\cite{lss} techniques), a Memory State-Space Model which consists of a Dynamic Memory Bank module to learn the temporal-aware latent dynamics and a Static Scene Propagation module to learn the spatial-aware latent statics, a Decoder to predict the actions and 3D occupancy, and a Task Prompt to condition the feature extraction for different tasks. 

\subsection{Memory State-Space Model}

\begin{figure}[t]
	\centering
	\includegraphics[width=0.48\textwidth]{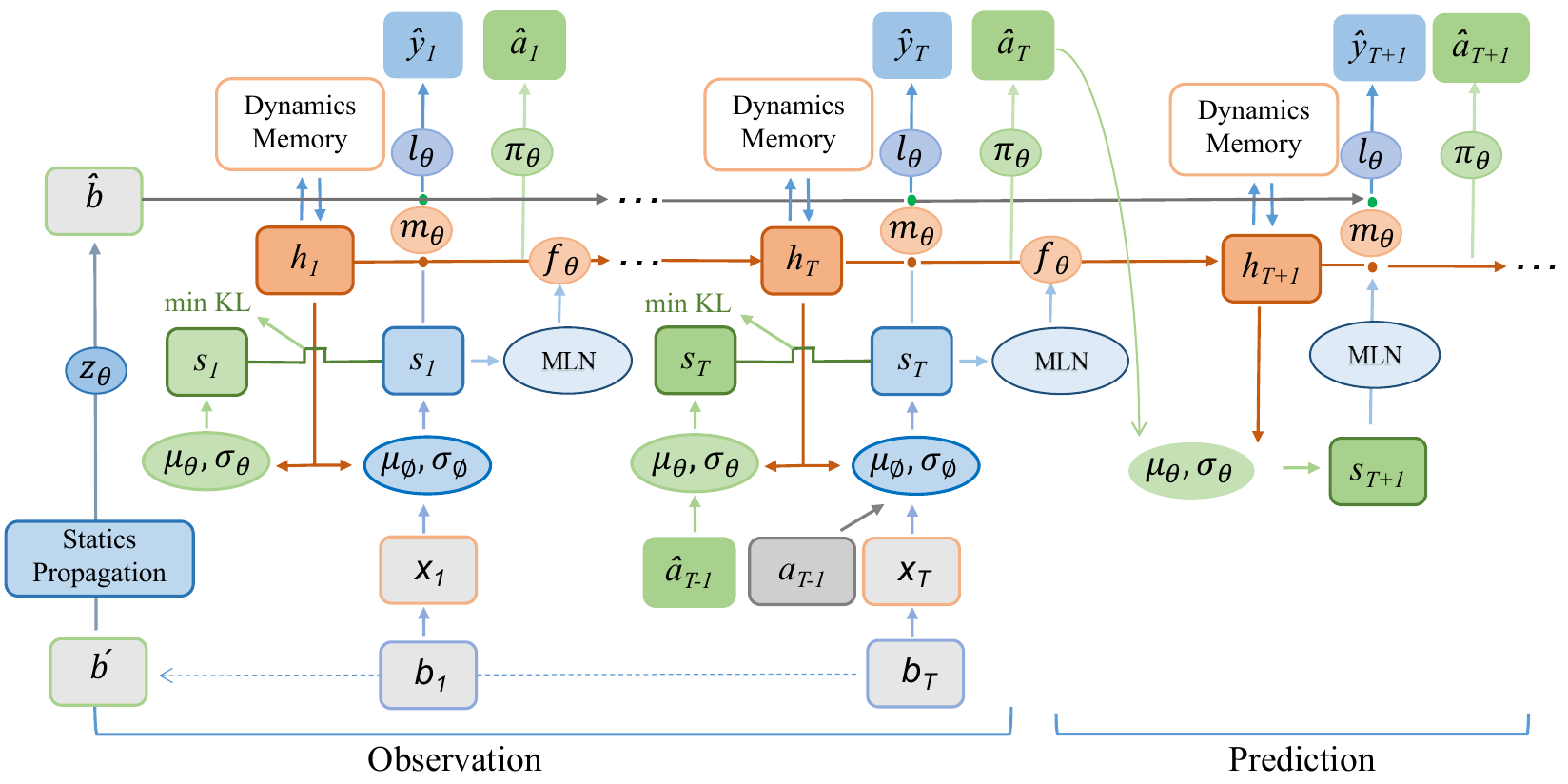}
	\caption{Overall architecture of proposed the Memory State-Sapce Model (MSSM). MSSM divides the transmitted information into two categories: temporal-aware information and spatial-aware information. The Dynamic Memory Bank module utilizes motion-aware layer normalization (MLN) to encode temporal-aware attributes and engages in information interaction with the dynamically updated memory bank. Meanwhile, the Static Scene Propagation module employs BEV features to represent spatial-aware latent statics, which are directly conveyed to the decoder.
	}
	\label{fig:mssm}
\end{figure}

As autonomous vehicle moves, it sequentially conveys two types of information within its observations: the temporal-aware information linked to alterations in the scene due to object mobility, and the spatial-aware information associated with scene context~\cite{contextwm}. As illustrated in Fig.~\ref{fig:mssm}, to address these dynamic agents and spatial scenes separately for 4D pre-training, we propose the Dynamic Memory Bank module for temporal-aware latent dynamics and the Static Scene Propagation module for spatial-aware latent statics. Next, we will begin by introducing the probabilistic model for temporal modelling, followed by detailed presentations of the Dynamic Memory Bank module and Static Scene Propagation module.

\paragraph{Probabilistic Modelling.}
To imbue the model with the capability for temporal modelling, we first introduce two latent variables $(h_{1:T}, s_{1:T})$, where $h_t$ represents the history and $s_t$ signifies the stochastic state. $h_t$ is updated with the past histories $h_{1:t-1}$ and stochastic states $s_{1:t-1}$. 

When images are observed, current scene perception can be obtained by utilizing past and current images. However, when predicting the future, in the absence of input images, we rely solely on past histories and states $(h_{1:t-1}, s_{1:t-1})$ to predict the future states. This predictive process is akin to the probabilistic generative models~\cite{vae}. For predicting future, we follow Recurrent State-Space Model~\cite{plas_wm} and construct both the posterior state distribution $q(s_t|o_{\leq t},a_{< t})$ and the prior state distribution $p(s_t|h_{t-1},s_{t-1})$. The objective is to match the prior distribution (the anticipated outcome based on past histories and states) with the posterior distribution (the outcome derived from observed multi-camera images and actions)~\cite{mile}.  

Considering the high dimensionality of BEV features, we transform them into a 1D vector $x_t\in \mathbb{R}^{512}$ and subsequently sample the Gaussian distribution from $(h_t, a_{t-1},x_t)$ to generate the posterior state distribution:

\begin{footnotesize} 
	\begin{equation} \label{post}
	q(s_t|o_{\leq t},a_{< t})\backsim \mathcal{N}(\mu _{\phi }(h_t,a_{t-1},x_t),\sigma _{\phi }(h_t, a_{t-1},x_t)\textbf{\textit{I}}),
	\end{equation} 
\end{footnotesize}
where $s_t$ is parameterised as a normal distribution with diagonal covariance and the initial distribution is set as $s_1\backsim \mathcal{N}(0,\textbf{\textit{I}})$. $(\mu _{\phi }, \sigma _{\phi })$ are multi-layer perceptrons that parametrise the posterior state distribution. 

In the absence of observed images, the model derives the prior state distribution based on historical information and predicted action:

\begin{footnotesize} 
	\begin{equation} \label{prior}
	p(s_t|h_{t-1},s_{t-1})\backsim \mathcal{N}(\mu _\theta (h_t,\hat{a}_{t-1}),\sigma_\theta (h_t,\hat{a}_{t-1})\textbf{\textit{I}}),
	\end{equation} 
\end{footnotesize}
where $(\mu _{\theta }, \sigma _{\theta })$ parameterizes the prior state distribution. $\pi _\theta$ is the policy network for predicting action $\hat{a}_{t-1}$ with past history $h_{t-1}$ and state $s_{t-1}$. Following MILE~\cite{mile}, we utilize MLP for action prediction, including velocity and steering.
\paragraph{Dynamic Memory Bank.}
In the process of temporal information propagation, we aim to account for the movement of objects by incorporating motion parameters.   Following StreamPETR~\cite{streampetr}, we introduce motion-aware layer normalization (MLN) into the latent dynamics propagation process. We define $K$ moving objects and estimate their velocities. The motion attributes consist of velocity $v$, and relative time interval $\Delta t$. $(v, \Delta t)$ are flattened and converted into affine vectors $\gamma $ and $\beta$ through two linear layers $(\xi_1, \xi_2)$: $\gamma = \xi_1(v, \Delta t),
\beta = \xi_2(v, \Delta t).$ 
Then an affine transformation is executed to yield the motion-aware latent stochastic state, denoted as $\tilde{  s}_{t} = \gamma \cdot LN(  s_{t}) + \beta.$
With the motion of the vehicle, the deterministic history $  h_t$ can establish a dynamic memory bank $  h_{1:t}$. The refined deterministic history $\tilde{  h}_t$ is obtained via the cross-attention mechanism with the dynamics memory bank.
The transition of deterministic history is set as $  h_{t+1}= f_\theta(\tilde{  h}_t, \tilde{  s}_{t}).$
\paragraph{Static Scene Propagation.} 
As the vehicle moves, consecutive frames of the scene typically depict minimal alterations, with a prominent presence of static objects such as roads, trees, and traffic signs constituting the scene's predominant content. Converting input images into a 1D vector would lead to the loss of crucial information. In addition to conveying temporal-aware information, the world model should also be able to model spatial-aware information.

As shown in Fig.~\ref{fig:mssm}, we randomly select a frame $o^{'}$ from frames 1 to $T$ and use its BEV features $b^{'}$ to construct a latent static representation $\hat{b}=z_\theta(b^{'})$ describing the spatio-aware structure. We combine the spatio-aware latent statics $\hat{b}$ and the temporal-aware latent dynamics $s_t$ in channel-wise manner. We opt not to use warping operations, allowing the model to learn a robust global representation of the entire scene and $s_t$ to focus on capturing motion information. As $s_t$ is learned from the BEV features $b_t$, during the model training process, BEV features simultaneously acquire representations for static scene and motion information. This holistic representation is subsequently utilized in the subsequent decoder network.

\subsection{3D Occupancy Prediction}
Aiming at a comprehensive understanding of surrounding scenes in autonomous driving, we model the physical world into the 3D occupancy structure, utilizing the geometric form of occupancy to depict the surrounding environment of the vehicle~\cite{mahjourian2022occupancy,occupancy,liu2023lidar,occnet,khurana2022differentiable}. In contrast to other world models that reconstruct the input 2D images~\cite{dreamerv2,gaia}, the 3D occupancy decoder can introduce geometric priors of the surrounding world through pre-training to vision-based models. Unlike depth estimation pre-training~\cite{dd3d,ppgeo}, which primarily represents object surfaces, 3D occupancy can represent the entire structure. Furthermore, unlike MILE's BEV segmentation target~\cite{mile}, which omits crucial height information, 3D occupancy provides a more comprehensive description of objects. The 3D occupancy decoder is set as $  \hat{y}_{t}= l_\theta(m_\theta(\tilde{  h}_t,   s_{t}), \hat{b}),$
where $m_\theta$ is the network expanding the 1D features to the dimensions of BEV, and $l_\theta$ is the 3D convolutional network for predicting occupancy.

Reconstructing 3D occupancy as the pre-text task has been demonstrated to be effective by pre-training algorithms like OccNet~\cite{occnet} and UniScene~\cite{uniscene}. In comparison to OccNet and UniScene, we further extend to 4D occupancy pre-training, introducing additional prior knowledge through spatio-temporal modelling.

\subsection{Task Prompt}
While the designed pre-text task through the world model enables the learning of spatio-temporal representations, different downstream tasks focus on distinct  information~\cite{wang2023tsp,liang2023visual}. For instance, the 3D object detection task emphasizes current spatio-aware information, while future prediction tasks prioritize temporal-aware information. Excessive focus on future information, such as the future position of a vehicle, could be detrimental to 3D object detection task.

To mitigate this problem, inspired by Semantic Prompt for few-shot image recognition~\cite{sp} and Visual Exemplar
driven Prompts for multi-task learning~\cite{liang2023visual}, we introduce the concept of ``Task Prompt'', providing specific cues to different heads to guide them in extracting the task-aware features. Acknowledging the semantic connections that exist among different tasks, we leverage the Large Language Model $g_\varphi (\cdotp )$ (\eg, BERT~\cite{bert}, CLIP~\cite{clip}) to construct these task prompts. For instance, the task prompt $p^{text}$ for the 3D occupancy reconstruction task that focuses on the current scene is set as straightforward as ``The task is to predict the 3D occupancy of the current scene''. We input the prompt $p^{text}$ into $g_\varphi (\cdotp)$ to acquire prompt encodings $g_\varphi (p^{text})$. Subsequently, we employ AdaptiveInstanceNorm~\cite{mile} and CNNs to expand it to the dimensions of BEV, denoted as $q_\varphi(g_\varphi (p^{text}))$, to integrate it with the learned spatio-temporal features.

\subsection{Pre-training Objective}
The pre-training objectives of DriveWorld involve minimizing the divergence between post and prior state distributions (\ie Kullback-Leibler (KL) divergence) and minimizing the loss related to past and future 3D occupancy (\ie Cross-Entropy loss (CE)) and actions (\ie L1 loss). 
We depicted the model observing inputs over $T$ timesteps, followed by envisioning future 3D occupancy and actions for $L$ steps. The overall loss function of DriveWorld is:

\begin{scriptsize }
\begin{equation} \label{loss}
\begin{aligned}
loss =&\displaystyle\sum_{t=1}^{T}[{\rm KL}(q(s_t|o_{\leq t,a_{<t}})\parallel p(s_t|h_{t-1},s_{t-1}))+{\rm CE}(\hat{y}_t,y_t)+\\
&{\rm L1}(\hat{a}_t,a_t)]
+\displaystyle\sum_{k=1}^{L}[{\rm CE}(\hat{y}_k,y_k)+{\rm L1}(\hat{a}_k,a_k)].
\end{aligned}
\end{equation}
\end{scriptsize }

For the OpenScene dataset~\cite{openscene}, we also utilize an L2 loss for occupancy flow prediction.
DriveWorld is based on the Probabilistic Generative Model~\cite{vae,dreamerv2,mile}.
For the detailed derivation of the loss function, please refer to Section~\ref{lbd} in the supplementary material. 

\subsection{Fine-tuning on Downstream Tasks}
Through DriveWorld, we acquire spatio-temporal BEV representations. Specifically, the network between image feature extraction and the generation of BEV features (\ie, encoder) is pre-trained. During fine-tuning, both the encoder and decoder (\ie head network for different tasks) with Task Prompts are trained simultaneously.

\section{Experiments}

\subsection{Experimental Setup}

\paragraph{Dataset.}  
We pre-train on the autonomous driving dataset nuScenes~\cite{nuscenes} and the largest-scale 3D occupancy dataset OpenScene~\cite{openscene}, and fine-tune on nuScenes. Evaluation settings are the same as UniAD~\cite{uniad}. For detailed dataset descriptions, please refer to the Section~\ref{data} in the supplemental material..
\paragraph{Pre-training.}
In alignment with BEVFormer~\cite{bevformer} and UniAD~\cite{uniad}, we employ ResNet101-DCN~\cite{resnet} as the foundational backbone. For 3D occupancy prediction, we establish a voxel size of $16\times200\times200$. The learning rate is set as 2$\times$10$^{-4}$. By default, the pre-training phase encompasses 24 epochs. The model observes inputs over $T$ = 4 steps, and the future prediction is set at $L$ = 4 steps.

\paragraph{Fine-tuning.}
In the fine-tuning stage, we retain the pre-trained encoder that generates BEV features and fine-tune downstream tasks. For the 3D detection task, we employed the BEVFormer~\cite{bevformer} framework, fine-tuning its parameters without freezing the encoder, and conducted training for 24 epochs. Regarding other autonomous driving tasks, we utilized the UniAD~\cite{uniad} framework and loaded our fine-tuned BEVFormer weights to UniAD, adhering to a standard 20-epoch training protocol for all tasks. For UniAD, we followed its experimental setup, which involved training for 6 epochs in stage 1 and 20 epochs in stage 2. Experiments are conducted with 8 NVIDIA Tesla A100 GPUs.

\subsection{Ablation Studies}
We first perform thorough ablation studies with UniAD~\cite{uniad} (only fine-tune on the Stage 1 with queue length of 3 for efficiency) pre-trained on nuScenes training set to validate the effectiveness of each component of DriveWorld. 
\paragraph{Component Analysis.}
We first validate the effectiveness of the proposed Memory State-Space Model (MSSM) module. As shown in Tab.~\ref{tab:ablation}, pre-training with the Recurrent State-Space Model (RSSM)~\cite{plas_wm} results in significantly poor 3D detection performance. This is attributed to RSSM having a 1D tensor for latent dynamics, which cannot effectively retain context information, consequently causing model disruption during pre-training. However, when Static Scene Propagation (SSP) is integrated into MSSM, direct reconstruction using BEV features leads to an approximately 1\% improvement in performance. Upon introducing Dynamic Memory Bank (DMB), performance drops in 3D detection and online mapping but improves in tracking. In motion prediction tasks, a wide perceptual field is likely necessary for the model to perform effectively. However, in detection tasks, precise localization is crucial, and a broad perceptual field could potentially introduce additional noise into the detection process. The subsequent introduction of Motion-aware Layer Normalization (MLN) yields improvements in all perception tasks. This demonstrates the importance of incorporating motion attributes when transferring dynamic information. Finally, the inclusion of the proposed task prompt decouples different information for distinct tasks, leading to further improvements in perception performance. 
\begin{table}[t]
	\centering
	\resizebox{0.5\textwidth}{!}
	{
		\begin{tabular}{c|ccc|c|cc|cc|cc}
			\toprule
			\multirow{2}*{\textbf{RSSM}}&\multicolumn{3}{c|}{\textbf{MSSM}}&\multirow{2}*{\textbf{Task Prompt}}&\multicolumn{2}{c|}{\textbf{Detection}} &\multicolumn{2}{c|}{\textbf{Tracking}} &\multicolumn{2}{c}{\textbf{Mapping}} \\
			&\textbf{SSP}&\textbf{DMB}&\textbf{MLN}&&\textbf{mAP}$\uparrow$&\textbf{NDS}$\uparrow$&\textbf{AMOTA}$\uparrow$&\textbf{AMOTP}$\downarrow$&\textbf{IoU-lane}$\uparrow$&\textbf{IoU-road}$\uparrow$ \\
			\midrule
			& & & &&0.416&0.517&0.355 &1.336 &0.301 &0.671 \\
			\midrule
			$\checkmark$& & & &&0.381&0.494&- &- &- &- \\
			\midrule
			& $\checkmark$&& &&0.429&0.528&0.365 &1.327 &0.319 &0.688\\
			\midrule
			&$\checkmark$ &$\checkmark$ & &&0.425&0.524&0.370 &1.320 &0.312 &0.686 \\
			\midrule
			&$\checkmark$ &$\checkmark$ &$\checkmark$ &&0.432&0.531&0.373 &1.312 &0.326 &0.698 \\
			\midrule
			\rowcolor{gray!10}&$\checkmark$ &$\checkmark$ &$\checkmark$ &$\checkmark$&\textbf{0.436}&\textbf{0.534}&\textbf{0.379} &\textbf{1.308} &\textbf{0.329} &\textbf{0.705} \\
			
			\bottomrule
		\end{tabular}
	}
	\caption{Ablation studies of each component of DriveWorld.}
	\label{tab:ablation}
\end{table}
\begin{table}[t] \tiny
	\centering
	\resizebox{0.5\textwidth}{!}
	{
		\begin{tabular}{c|c|cc|cc}
			\toprule
			\multirow{2}*{\textbf{Pre-train }}&\multirow{2}*{\textbf{Fine-tune }}&\multicolumn{2}{c|}{\textbf{Detection}} &\multicolumn{2}{c}{\textbf{Tracking}}  \\
			&&\textbf{mAP}$\uparrow$&\textbf{NDS}$\uparrow$&\textbf{AMOTA}$\uparrow$&\textbf{AMOTP}$\downarrow$ \\
			\midrule
			0\%&100\%&0.416 & 0.517&0.355 &1.336 \\	
			\midrule
			50\%&100\%&0.425 &0.523 &0.364 &1.323 \\
			\midrule
			\rowcolor{gray!10}100\%&75\%&0.418 & 0.518&0.358 &1.331\\
			\midrule
			100\%&100\%&\textbf{0.436} &\textbf{0.534} &\textbf{0.379} &\textbf{1.308} \\
			
			\bottomrule
		\end{tabular}
	}
	\caption{Ablation studies of different scales of dataset.}
	\label{tab:ablation_scale}
\end{table}
\begin{table}[t]
	\centering
	\resizebox{0.5\textwidth}{!}
	{
		{
			\begin{tabular}{c|c|c|c|c|c|c|c}
				\toprule
				\textbf{Method} &\textbf{mAP}$\uparrow$&\textbf{NDS}$\uparrow$&\textbf{mATE}$\downarrow$&\textbf{mASE}$\downarrow$&\textbf{mAOE}$\downarrow$&\textbf{mAVE}$\downarrow$&\textbf{mAAE}$\downarrow$ \\
				\midrule
				DETR3D~\cite{detr3d} &0.349&0.434&0.716&0.268&0.379&0.842&0.200\\
				UVTR~\cite{uvtr}&0.379 &0.483 &0.731 &0.267 &0.350 &0.510 &0.200\\
				\midrule
				BEVFormer$^{*}$~\cite{bevformer} &0.377&0.477&0.708&0.280&0.450&0.433&0.198\\
				+ FCOS3D~\cite{fcos3d} &0.416&0.517&0.673&0.274&0.372&0.394&0.198\\
				+ OccNet~\cite{occnet} &0.436&0.532&0.655&0.273 &0.372 &0.349 &0.182\\
				+ UniScene~\cite{uniscene} &0.438&0.534&0.656&0.271 &0.371 &0.348 &0.183\\
				+ BEVDistill~\cite{bevdistill} &0.439&0.536&0.653&0.271 &0.372 &0.343 &0.180\\
				\midrule
				\rowcolor{gray!10}+ \textbf{DriveWorld}$^{\dagger }$ &0.442$^{\textcolor{teal} {+6.5\%}}$&0.536$^{\textcolor{teal}{+5.9\%}}$&0.650&0.268 &0.370 &0.342 &0.183 \\
				\rowcolor{gray!10}+ \textbf{DriveWorld}$^{\ddagger }$ &$\textbf{0.452}^{\textcolor{teal} {+7.5\%}}$&$\textbf{0.545}^{\textcolor{teal} {+6.8\%}}$&\textbf{0.642}&\textbf{0.264}&\textbf{0.359}&\textbf{0.324}&\textbf{0.176}\\
				
				\bottomrule
			\end{tabular}
	}}
	\caption{Quantitative 3D object detection performance. ${*}$: we retrain BEVFormer~\cite{bevformer} with 2D ImageNet pre-training~\cite{resnet}.}
	\label{tab:detection}
\end{table}

\begin{table}[t]
	\centering
	\resizebox{0.5\textwidth}{!}
	{
		\begin{tabular}{c|c|c|c|c}
			\toprule
			\textbf{Method}&\textbf{Lanes}$\uparrow$&\textbf{Drivable}$\uparrow$&\textbf{Divider}$\uparrow$&\textbf{Crossing}$\uparrow$\\
			\midrule
			BEVFormer~\cite{bevformer}&23.9&\bf77.5&-&- \\
			BEVerse~\cite{beverse} &-&-&\bf30.6&\bf17.2 \\
			\midrule
			UniAD~\cite{uniad}&31.3&69.1&25.7&13.8 \\
			+ OccNet~\cite{occnet}&32.1&70.2&26.3&14.2 \\ 
			+ UniScene~\cite{uniscene}&32.5&70.5&26.9&14.9 \\ 
			+ BEVDistill~\cite{bevdistill}&32.7&70.4&26.8&14.7 \\ 
			\midrule
			\rowcolor{gray!10}+ \textbf{DriveWorld}$^{\dagger }$  &33.4$^{\textcolor{teal} {+2.1\%}}$&71.3$^{\textcolor{teal} {+2.2\%}}$&27.9$^{\textcolor{teal} {+2.2\%}}$&15.2$^{\textcolor{teal} {+1.4\%}}$\\
			\rowcolor{gray!10}+ \textbf{DriveWorld}$^{\ddagger }$ &\bf34.2$^{\textcolor{teal} {+2.9\%}}$&73.7$^{\textcolor{teal} {+4.6\%}}$&29.5$^{\textcolor{teal} {+3.8\%}}$&\bf17.2$^{\textcolor{teal} {+3.4\%}}$\\
			\bottomrule
		\end{tabular}
	}
	\caption{Quantitative online mapping performance.}
	\label{tab:mapping}
\end{table}
\paragraph{Dataset Scale.}

We also investigate the influence of pre-training and fine-tuning data volumes. Tab.~\ref{tab:ablation_scale} illustrates that augmenting the volume of data used in pre-training leads to improved performance in downstream tasks. Importantly, using just 75\% of the data for fine-tuning still results in comparable performance. This finding underscores the efficacy of our 4D pre-training approach in reducing the data requirements by 25\%, which translates into considerable cost savings in terms of annotation and, as such, represents a substantial practical and economic advantage.

\subsection{Main Results}

In this section, we validate the effectiveness of our proposed 4D pre-training approach based on the world model across various autonomous driving tasks. In addition to comparing it with state-of-the-art autonomous driving algorithms, we also contrast it with various pre-training algorithms, including 2D ImageNet pre-training~\cite{resnet}, monocular 3D detection algorithm FCOS3D~\cite{fcos3d}, knowledge distillation algorithm BEVDistill~\cite{bevdistill}, and 3D occupancy pre-training algorithms OccNet~\cite{occnet} and UniScene~\cite{uniscene}, to provide a comprehensive assessment. The symbol $^{\dagger}$ denotes pre-training with the training set of nuScenes~\cite{nuscenes}, while $^{\ddagger}$ signifies pre-training with the training set of OpenScene~\cite{openscene}. For fine-tuning, we utilize the same decoder head as UniAD~\cite{uniad}. ``+X'' indicates experimental results obtained after fine-tuning UniAD with different pre-trained model X.

\paragraph{3D Object Detection.}

We first evaluate the performance of the multi-camera 3D object detection task. The results presented in Tab.~\ref{tab:detection} show that our 4D pre-training approach based on the world model, as opposed to BEVFormer relying solely on 2D ImageNet~\cite{resnet} pre-training, delivers a substantial increase of 7.5\% in mAP and 6.8\% in NDS. BEVFormer with FCOS3D pre-training, specifically tailored for monocular 3D object detection, outperforms models that rely solely on 2D pre-training resulting in a commendable 4\% increase in performance. OccNet, UniScene, and BEVDistill, which leverage 3D occupancy reconstruction and knowledge distillation as the pre-training target, result in an additional 2\% performance increase.
These findings underscore the effectiveness of 3D pre-training when compared to traditional 2D pre-training paradigms. Our innovative DriveWorld, which introduces 4D spatio-temporal pre-training, exhibits a modest performance improvement over OccNet, UniScene, and BEVDistill on the nuScenes dataset. When extended to the large-scale occupancy dataset OpenScene for pre-training, it contributes to an additional 1\% performance enhancement. 
\paragraph{Online Mapping.}	
\begin{table}[t]
	\centering
	\resizebox{0.5\textwidth}{!}
	{
		\begin{tabular}{c|c|c|c|c}
			\toprule
			\textbf{Method}&\textbf{AMOTA}$\uparrow$&\textbf{AMOTP}$\downarrow$&\textbf{Recall}$\uparrow$&\textbf{IDS}$\downarrow$\\
			\midrule
			QD3DT~\cite{qd3dt} &0.242&1.518&0.399&- \\
			MUTR3D~\cite{mutr3d} &0.294&1.498&0.427&3822 \\
			\midrule
			UniAD~\cite{uniad} &0.359&1.320&0.467&906 \\
			+ OccNet~\cite{occnet} &0.363&1.315&0.474&950 \\
			+ UniScene~\cite{uniscene} &0.373&1.312&0.484&832 \\
			+ BEVDistill~\cite{bevdistill} &0.376&1.310&0.489&812 \\
			\midrule
			\rowcolor{gray!10}+ \textbf{DriveWorld}$^{\dagger }$    &0.385$^{\textcolor{teal} {+2.6\%}}$&1.303$^{\textcolor{teal} {-1.7\%}}$&0.511$^{\textcolor{teal} {+4.4\%}}$&710$^{\textcolor{teal} {-196}}$ \\
			\rowcolor{gray!10}+ \textbf{DriveWorld}$^{\ddagger }$  &\bf0.412$^{\textcolor{teal} {+5.3\%}}$&\bf1.266$^{\textcolor{teal} {-5.4\%}}$&\bf0.545$^{\textcolor{teal} {+7.8\%}}$&\bf701$^{\textcolor{teal} {-205}}$\\
			\bottomrule
		\end{tabular}
	}
	\caption{Quantitative multi-object tracking performance.}
	\label{tab:mot}
\end{table}

\begin{table}[t]
	\centering
	\resizebox{0.5\textwidth}{!}
	{
		\begin{tabular}{c|c|c|c|c}
			\toprule
			\textbf{Method}&\textbf{minADE(m)}$\downarrow$&\textbf{minFDE(m)}$\downarrow$&\textbf{MR}$\downarrow$&\textbf{EPA}$\uparrow$\\
			\midrule
			PnPNet~\cite{pnpnet} &1.15&1.95&0.226&0.222\\
			ViP3D~\cite{vip3d} &2.05&2.84&0.246&0.226\\
			\midrule
			UniAD~\cite{uniad}&0.71&1.02&0.151&0.456 \\
			+ OccNet~\cite{occnet}&0.70&1.02&0.146&0.459 \\
			+ UniScene~\cite{uniscene}&0.69&1.01&0.148&0.457 \\
			+ BEVDistill~\cite{bevdistill}&0.70&0.99&0.146&0.460 \\
			\midrule
			\rowcolor{gray!10}+ \textbf{DriveWorld}$^{\dagger }$  &0.67$^{\textcolor{teal} {-0.04}}$&0.94$^{\textcolor{teal} {-0.08}}$&0.140$^{\textcolor{teal} {-0.011}}$&0.468$^{\textcolor{teal} {+0.012}}$\\
			\rowcolor{gray!10}+ \textbf{DriveWorld}$^{\ddagger }$ &\bf0.61$^{\textcolor{teal} {-0.10}}$&\bf0.91$^{\textcolor{teal} {-0.11}}$&\bf0.136$^{\textcolor{teal} {-0.025}}$&\bf0.503$^{\textcolor{teal} {+0.047}}$\\ 
			\bottomrule
		\end{tabular}
	}
	\caption{Quantitative motion forecasting performance.}
	\label{tab:forecasting}
\end{table}
\begin{table}[t]
	\centering
	\resizebox{0.5\textwidth}{!}
	{
		\begin{tabular}{c|c|c|c|c}
			\toprule
			\textbf{Method}&\textbf{IoU-n}$\uparrow$&\textbf{IoU-f}$\uparrow$&\textbf{VPQ-n}$\uparrow$&\textbf{VPQ-f}$\uparrow$\\
			\midrule
			ST-P3~\cite{stp3} &-&38.9&-&32.1\\
			BEVerse~\cite{beverse} &61.4&40.9&54.3&36.1\\
			\midrule
			UniAD~\cite{uniad} &63.4&40.2&54.7&33.5\\
			+ OccNet~\cite{occnet} &63.9&40.8&55.1&34.2\\
			+ UniScene~\cite{uniscene} &64.3&41.2&55.3&34.9\\
			+ BEVDistill~\cite{bevdistill} &64.1&40.9&54.9&33.8\\
			\midrule
			\rowcolor{gray!10}+ \textbf{DriveWorld}$^{\dagger }$  &65.3$^{\textcolor{teal} {+1.9\%}}$&42.4$^{\textcolor{teal} {+2.2\%}}$&56.7$^{\textcolor{teal} {+2.0\%}}$&35.3$^{\textcolor{teal} {+1.8\%}}$\\
			\rowcolor{gray!10}+ \textbf{DriveWorld}$^{\ddagger }$ &\bf66.2$^{\textcolor{teal} {+2.8\%}}$&\bf45.2$^{\textcolor{teal} {+5.0\%}}$&\bf58.1$^{\textcolor{teal} {+3.4\%}}$&\bf36.9$^{\textcolor{teal} {+3.4\%}}$\\
			\bottomrule
		\end{tabular}
	}
	\caption{Quantitative occupancy prediction performance.}
	\label{tab:occ}
\end{table}

\begin{table}[t]
	\centering
	\resizebox{0.5\textwidth}{!}
	{
		\begin{tabular}{c|cccc|cccc}
			\toprule
			\multirow{2}*{\textbf{Method}} &\multicolumn{4}{c|}{\textbf{L2(m)$\downarrow$}} &\multicolumn{4}{c}{\textbf{Col.Rate(\%)$\downarrow$}}\\
			&\textbf{1s} &\textbf{2s} &\textbf{3s} &\textbf{Avg.} &\textbf{1s}&\textbf{2s} &\textbf{3s} &\textbf{Avg.}\\
			\midrule
			ST-P3~\cite{stp3}  &1.33&2.11&2.90&2.11&0.23&0.62&1.27&0.71\\
			BEVGPT~\cite{bevgpt}  &0.39 &0.88 &1.70 &1.22&-&-&-&-\\
			\midrule
			UniAD~\cite{uniad} &0.48&0.96&1.65&1.03&0.05&0.17&0.71&0.31\\
			+ OccNet~\cite{occnet} &0.49&0.95&1.64&1.02&0.07&0.15&0.69&0.30\\
			+ UniScene~\cite{uniscene} &0.47&0.91&1.56&0.98&0.05&0.16&0.64&0.28
			\\
			+ BEVDistill~\cite{bevdistill} &0.46&0.92&1.60&0.99&0.05&0.16&0.67&0.29
			\\
			\midrule
			\rowcolor{gray!10}+ \textbf{DriveWorld}$^{\dagger }$  &0.47$^{\textcolor{teal} {-0.01}}$&0.86$^{\textcolor{teal} {-0.10}}$&1.42$^{\textcolor{teal} {-0.23}}$&0.92$^{\textcolor{teal} {-0.11}}$&0.05&0.13$^{\textcolor{teal} {-0.04}}$&0.59$^{\textcolor{teal} {-0.12}}$&0.26$^{\textcolor{teal} {-0.05}}$\\
			\rowcolor{gray!10}+ \textbf{DriveWorld}$^{\ddagger }$  &\bf0.34$^{\textcolor{teal} {-0.14}}$&\bf0.67$^{\textcolor{teal} {-0.29}}$&\bf1.07$^{\textcolor{teal} {-0.58}}$&\bf0.69$^{\textcolor{teal} {-0.34}}$&\bf0.04$^{\textcolor{teal} {-0.01}}$&\bf0.12$^{\textcolor{teal} {-0.05}}$&\bf0.41$^{\textcolor{teal} {-0.30}}$&\bf0.19$^{\textcolor{teal} {-0.12}}$\\
			\bottomrule
		\end{tabular}
	}
	\caption{Quantitative planning performance.}
	\label{tab:planning}
\end{table}

\begin{figure}[t]
	\centering
	\includegraphics[width=0.5\textwidth]{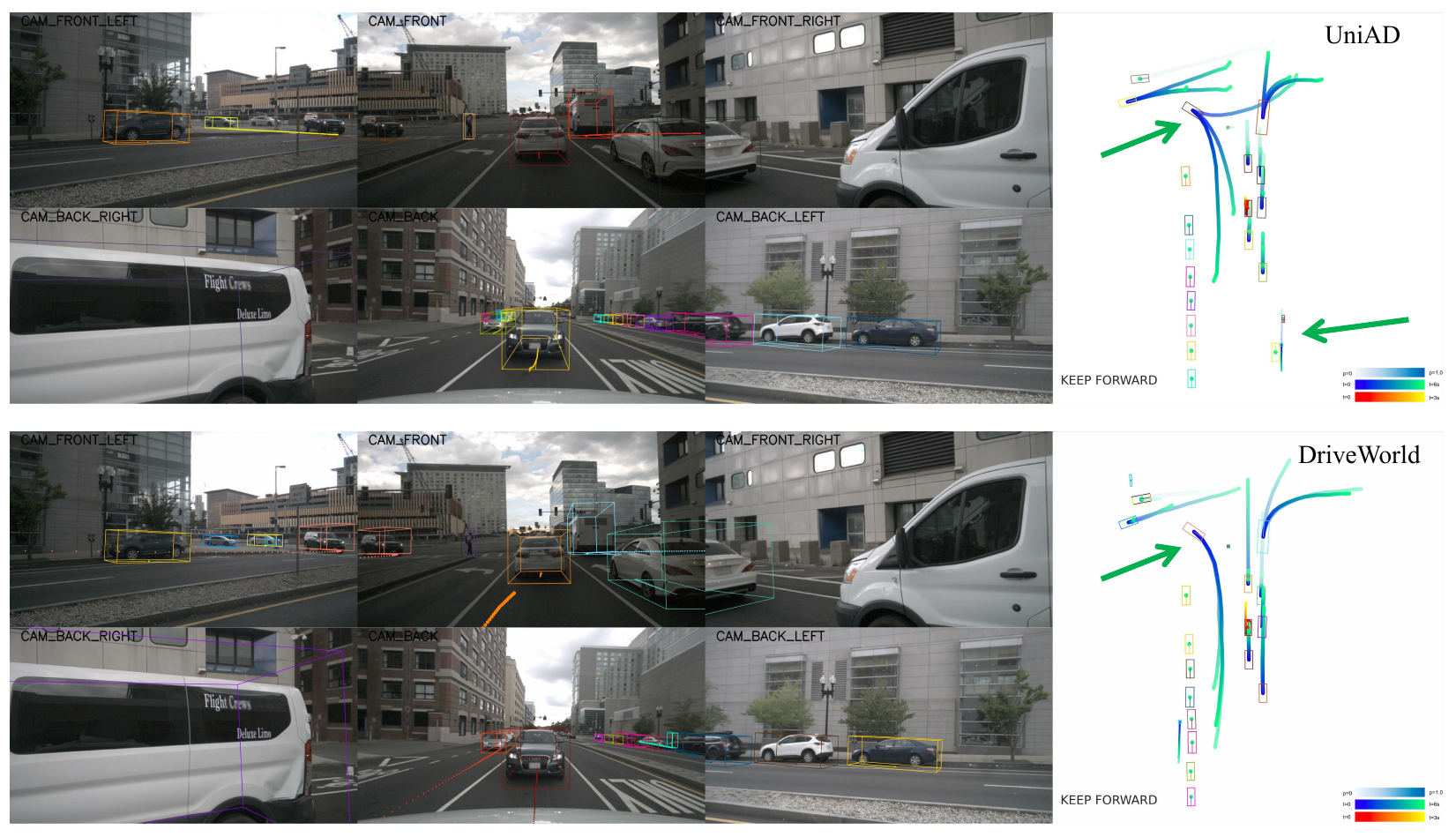} 
	\caption{Visual comparison between UniAD~\cite{uniad} (Top) and our DriveWorld (Bottom).}
	\label{fig:compare}
\end{figure}
We validate the performance on the online mapping task. As shown in Tab.~\ref{tab:mapping}, compared to UniAD, pre-training with 3D occupancy in OccNet and UniScene results in an improvement of about 1\% in IoU, and the knowledge distillation algorithm BEVDistill also enhances performance by 1\%. After our 4D pre-training on nuScenes, there is a 2\% improvement and a 3\% improvement after pre-training on OpenScene.
\paragraph{Multi-object Tracking.}
We further evaluate the performance on the multi-object tracking task, which demands a deeper consideration of temporal information. Tab.~\ref{tab:mot} illustrates the outcomes of this evaluation. It is evident that leveraging DriveWorld's 4D pre-training results in a notable enhancement of 2.6\% in terms of AMOTA. Impressively, after pre-training on OpenScene, this performance boost becomes even more significant, reaching a substantial 5.3\% increase in AMOTA. In contrast, pre-training with OccNet, UniScene, and BEVDistill only provides a moderate improvement of 1.0\% in AMOTA. Furthermore, DriveWorld exhibits the lowest ID switch score, indicating that DriveWorld enables the model to consistently demonstrate temporal coherence for each tracklet.
\paragraph{Motion Forecasting.}
In the motion prediction task, as demonstrated in Tab.~\ref{tab:forecasting}, the improvement obtained from 3D pre-training (\eg, OccNet, UniScene, and BEVDistill) is notably limited. In contrast, our 4D pre-training approach, which encompasses the capability to forecast future states, significantly enhances the performance of the motion prediction task. Pre-training on nuScenes results in a reduction of 0.04m in minADE, while pre-training on OpenScene leads to a remarkable 0.1m decrease in minADE. This notable improvement is partly attributed to the larger data scale of OpenScene and the presence of valuable flow information in this dataset. 
\paragraph{Occupancy Prediction.}
The UniAD's occupancy prediction task is carried out in the 2D BEV view. As shown in Tab.~\ref{tab:occ}, after undergoing 4D occupancy pre-training on OpenScene, our model exhibits impressive enhancements: a 2.8\% increase in IoU-near, a 5\% boost in IoU-far, a 3.4\% gain in VPQ-near, and a 3.4\% rise in VPQ-far. This outcome underscores the effectiveness of our pre-training approach in achieving a more comprehensive reconstruction of 4D scenes.
\paragraph{Planning.}
We finally validate the effectiveness of the proposed 4D pre-training algorithm on the planning task. As illustrated in Tab.~\ref{tab:planning}, DriveWorld stands out by achieving new state-of-the-art planning results, reducing an 0.34m average L2 error and an average Collision rate of 0.12. These results surpass the prior best model, UniAD. UniAD integrates perception, prediction, and planning in a sequential fashion. Our 4D pre-training approach, which comprehensively reconstructs the 3D scene, enhances tasks focused on the current scene, such as detection and segmentation. and predicts future scenarios elevating tracking and forecasting capabilities. By combining these advantages, we further improve the performance of the final planning step. Consequently, we have developed a robust fundamental model for autonomous driving. 

\subsection{Qualitative Results}
The qualitative comparison between UniAD and DriveWorld is visualized in Fig.~\ref{fig:compare}. UniAD exhibited false positives in detecting distant objects, and the detection accuracy was improved by DriveWorld. Additionally, UniAD made trajectory prediction errors for turning vehicles, which was addressed by DriveWorld after 4D pre-training, allowing for accurate predictions of future changes.
\section{Conclusion}
We introduce DriveWorld, a world model-based 4D pre-training method for vision-centric autonomous driving. DriveWorld learns compact spatio-temporal BEV representations via a world model that predicts 3D occupancy based on the past multi-camera images and actions. We design a Memory State-Space Model for spatio-temporal modelling, employing a Dynamic Memory Bank module to learn temporal-aware representations and a Static Scene Propagation module to learn spatial-aware representations. Additionally, a Task Prompt is introduced to guide the model toward adaptively acquiring task-specific representations. Extensive experiments demonstrate that DriveWorld significantly enhances the performance of various autonomous driving tasks. 
The power of DriveWorld to represent 4D world knowledge opens new pathways for innovation within autonomous driving.
\paragraph{Limitations and Future Work.}
Currently, the annotation of DriveWorld is still based on LiDAR point clouds. It is essential to explore self-supervised learning for vision-centric pre-training. Besides, the effectiveness of DriveWorld has only been validated on the lightweight ResNet101 backbone; it is worthwhile to consider scaling up the dataset and the backbone size. We hope the proposed 4D pre-training method can contribute to the development of the foundation model for autonomous driving.  
{
    \small
    \bibliographystyle{ieeenat_fullname}
    \bibliography{main}
}
\clearpage
\maketitlesupplementary

\section{Pre-training Objective}
\label{lbd}
The proposed DriveWorld for 4D driving pre-training encompasses the following five components:

\begin{footnotesize} 
	\begin{equation} \label{world_model}
	\begin{aligned}
	{\rm BEV \, Representation \, Model:}\, &b_t\sim q_\phi  (b_t\mid o_t) \\
	{\rm Stochastic \, State \, Model:}\, &s_t\sim q_\phi  (s_t\mid h_{t},a_{t-1},o_t) \\
	{\rm Dynamic \, Transition \, Model:}\, &s_t\sim p_\theta (s_t\mid h_t,\hat{a}_{t-1}) \\
	{\rm Static \, Propagation \, Model:}\, &\hat{b}\sim p_\theta (\hat{b}\mid b^{'}) \\
	{\rm Action \, Decoder:}\, &\hat{a}_t\sim p_\theta  (\hat{a}_t\mid h_{t},s_t) \\
	{\rm 3D \, Occupancy \, Decoder:}\, &\hat{y}_t\sim p_\theta  (\hat{y}_t\mid h_{t},s_t,\hat{b}). \\
	\end{aligned}
	\end{equation} 
\end{footnotesize}

The joint probability distribution for DriveWorld is:

\begin{footnotesize} 
	\begin{equation} \label{v1}
	\begin{aligned}
	&p(h_{1:T},s_{1:T},y_{1:T+L},a_{1:T+L})=\\&\displaystyle\prod_{t=1}^{T}p(h_t,s_t|h_{t-1},s_{t-1},a_{t-1})p(y_t,a_t|h_t,s_t,\hat{b})\\
	&\displaystyle\prod_{k=1}^{L}p(h_k,s_k|h_{T},s_{T},a_{k-1})p(y_k,a_k|h_T,s_T,\hat{b}),
	\end{aligned}
	\end{equation} 
\end{footnotesize}

with 

\begin{footnotesize} 
	\begin{equation} \label{v21}
	p(h_t,s_t|h_{t-1},s_{t-1},a_{t-1})=p(h_t|h_{t-1},s_{t-1})p(s_t|h_t,a_{t-1}),
	\end{equation} 
\end{footnotesize}

\begin{footnotesize} 
	\begin{equation} \label{v22}
	p(y_t,a_t|h_{t},s_{t})=p(y_t|h_{t},s_{t},\hat{b})p(a_t|h_t,s_{t}),
	\end{equation}
\end{footnotesize} 

\begin{footnotesize} 
	\begin{equation} \label{v23}
	p(y_k,a_k|h_{T},s_{T})=p(y_k|h_{T},s_{T},\hat{b})p(a_{k}|h_T,s_{T}).
	\end{equation}
\end{footnotesize} 

Given that $h_t$ is deterministic~\cite{plas_wm,dreamerv2,mile}, we have $p(h_t|h_{t-1},s_{t-1})=\delta(h_t-f_\theta(\hat{h}_{t-1},\rm MLN(s_{t-1})))$. Consequently, to maximize the marginal likelihood of $p(y_{1:{T+L}},a_{1:{T+L}})$, it is imperative to infer the latent variables $s_{1:T}$. This is achieved through deep variational inference, wherein we introduce a variational distribution $q_{H,S}$ defined and factorized as follows:

\begin{footnotesize} 
	\begin{equation} \label{v3}
	\begin{aligned}
	q_{H,S}&\triangleq q(h_{1:T},s_{1:T}|o_{1:T},y_{1:T+L},a_{1:T+L})\\
	&=\displaystyle\prod_{t=1}^{T}q(h_t|h_{t-1},s_{t-1})q(s_t|o_{\le t},a_{< t}).
	\end{aligned}
	\end{equation} 
\end{footnotesize}

We parameterise this variational distribution with a neural network with weights $\phi $. By formalizing the above process as a generative probabilistic model, we can obtain a variational lower bound on the log evidence:

\begin{footnotesize}
	\begin{equation} \label{vae}
	\begin{aligned}
	&{\rm log} p(y_{1:T+L},a_{1:T+L})\ge \mathcal{L}(y_{1:T+L},a_{1:T+L};\theta ,\phi ) \\ &\triangleq  \displaystyle\sum_{t=1}^{T}\mathbb{E}_{h_{1:t},s_{1:t}\backsim q(h_{1:t},s_{1:t}|o_{\leq t},a_{<t})}[\underbrace{\log p(y_t|h_t,s_t,\hat{b})}_{{\rm past \ occupancy\, loss}}\\
	&\qquad+\underbrace{\log p(a_t|h_t,s_t)}_{{\rm past \ action\, loss}}]  +\displaystyle\sum_{k=1}^{L}\mathbb{E}_{h_{T},s_{T}\backsim q(h_{T},s_{T}|o_{\leq T},a_{<T})}\\
	&\qquad[\underbrace{\log p(y_{T+k}|h_T,s_T,\hat{b})}_{{\rm future \ occupancy\, loss}}+\underbrace{\log p(a_{T+k}|h_T,s_T)}_{{\rm future \ action\, loss}}] \\
	&\quad-\displaystyle\sum_{t=1}^{T}\mathbb{E}_{h_{1:t-1},s_{1:t-1}\backsim q(h_{1:t-1},s_{1:t-1}|o_{\leq t-1},a_{<t-1} )}\\
	&\qquad[\underbrace{D_{KL}(q(s_t|o_{\leq t},a_{<t})\parallel p(s_t|h_{t-1},s_{t-1}))}_{{\rm posterior\, and\, prior\, matching\, KL\, loss}}].
	\end{aligned}
	\end{equation} 
\end{footnotesize}

In Eqn.~\ref{vae}, we model $q(s_t|o_{1:t}, a_{1:t-1})$ as a Gaussian distribution, allowing for the closed-form computation of the Kullback-Leibler (KL) divergence. The modelling of actions as a Laplace distribution and 3D occupancy labels as a categorical distribution results in L1 and cross-entropy losses, respectively. 
\section{Lower Bound Derivation} 

Next, we will derive the variational lower bound in Eqn.~\ref{vae}. Let $q_{H,S}\triangleq q(h_{1:T},s_{1:T}|o_{1:T},y_{1:T+L},a_{1:T+L})$ be the variational distribution and $p(h_{1:T},s_{1:T}|a_{1:T+L},y_{1:T+L})$ be the posterior distribution. The Kullback-Leibler divergence between these two distributions is:

\begin{footnotesize} 
	\begin{equation} \label{1}
	\begin{aligned}
	&D_{KL}(q(h_{1:T},s_{1:T}|o_{1:T},y_{1:T+L},a_{1:T+L})  \\
	&\quad \parallel p(h_{1:T},s_{1:T}|y_{1:T+L},a_{1:T+L}))\\
	&=\mathbb{E}_{h_{1:T},s_{1:T}\backsim q_{H,S}}[{\log}\tfrac{q(h_{1:T},s_{1:T}|o_{1:T},y_{1:T+L},a_{1:T+L})}{p(h_{1:T},s_{1:T}|y_{1:T+L},a_{1:T+L})}]\\
	&=\mathbb{E}_{h_{1:T},s_{1:T}\backsim q_{H,S}}\\
	&\quad[{\log}\tfrac{q(h_{1:T},s_{1:T}|o_{1:T},y_{1:T+L},a_{1:T+L})p(y_{1:T+L},a_{1:T+L})}{p(h_{1:T},s_{1:T})p(y_{1:T+L},a_{1:T+L}|h_{1:T},s_{1:T})}]\\  
	&=\log p(y_{1:T+L},a_{1:T+L})- \\
	&\quad \mathbb{E}_{h_{1:T},s_{1:T}\backsim q_{H,S}}[\log p(y_{1:T+L},a_{1:T+L}|h_{1:T},s_{1:T})]+ \\ 
	&\quad D_{KL}(q(h_{1:T},s_{1:T}|o_{1:T},y_{1:T+L},a_{1:T+L})\parallel p(h_{1:T},s_{1:T})).
	\end{aligned}
	\end{equation} 
\end{footnotesize}

Since $D_{KL}(q(h_{1:T},s_{1:T}|o_{1:T},y_{1:T+L},a_{1:T+L})\parallel p(h_{1:T},s_{1:T}|y_{1:T+L},a_{1:T+L}))\geq 0$, we derive the following evidence lower bound:

\begin{footnotesize} 
	\begin{equation} \label{3}
	\begin{aligned}
	&\log p(y_{1:T+L},a_{1:T+L}) \geq\\ &\mathbb{E}_{h_{1:T},s_{1:T}\backsim q_{H,S}}[\log p(y_{1:T+L},a_{1:T+L}|h_{1:T},s_{1:T})] \\
	&- D_{KL}(q(h_{1:T},s_{1:T}|o_{1:T},y_{1:T+L},a_{1:T+L})\parallel p(h_{1:T},s_{1:T})).\\  
	\end{aligned}
	\end{equation} 
\end{footnotesize}

The two terms of this lower bound can be calculated separately. Firstly:

\begin{footnotesize} 
	\begin{equation} \label{4}
	\begin{aligned}
	&\mathbb{E}_{h_{1:T},s_{1:T}\backsim q_{H,S}}[\log p(y_{1:T+L},a_{1:T+L}|h_{1:T},s_{1:T})]\\ &=\mathbb{E}_{h_{1:T},h_{1:T}\backsim q_{H,S}}[\log\displaystyle\prod_{t=1}^{T}p(y_t|h_t,s_t,\hat{b})p(a_t|h_t,s_t)\\
	&\quad \displaystyle\prod_{k=1}^{L}p(y_k|h_T,s_T,\hat{b})p(a_k|h_T,s_T)] \\
	&=\displaystyle\sum_{t=1}^{T}\mathbb{E}_{h_{1:t},s_{1:t}\backsim q(h_{1:t},s_{1:t}|o_{\leq t},a_{<t})}[\log p(y_t|h_t,s_t,\hat{b})p(a_t|h_t,s_t)]\\   
	&\quad  +  \displaystyle\sum_{k=1}^{L}\mathbb{E}_{h_{T},s_{T}\backsim q(h_{T},s_{T}|o_{\leq t},a_{<t})} \\
	&\qquad[\log p(y_{T+k}|h_T,s_T,\hat{b})p(a_{T+k}|h_T,s_T)].\\  
	\end{aligned}
	\end{equation} 
\end{footnotesize}

Secondly, with $q(h_t|h_{t-1},s_{t-1})=p(h_t|h_{t-1},s_{t-1})$, we obtain:

\begin{footnotesize} 
	\begin{equation} \label{5}
	\begin{aligned}
	&D_{KL}(q(h_{1:T},s_{1:T}|o_{1:T},y_{1:T+L},a_{1:T+L})\parallel p(h_{1:T},s_{1:T}))\\
	&=D_{KL}(q(h_{1:T},s_{1:T}|o_{1:T},a_{1:T-1})\parallel p(h_{1:T},s_{1:T}))\\
	&=\int_{{h_{1:T},s_{1:T}}}q(h_{1:T},s_{1:T}|o_{1:T},a_{1:T-1}) \\
	&\quad\log \frac{q(h_{1:T},s_{1:T}|o_{1:T},a_{1:T-1})}{p(h_{1:T},s_{1:T})} d h_{1:T} d s_{1:T} \\
	&=\int_{{h_{1:T},s_{1:T}}}q(h_{1:T},s_{1:T}|o_{1:T},a_{1:T-1}) \\
	&\quad \log [\displaystyle\prod_{t=1}^{T}\frac{q(h_{t}|h_{t-1},s_{t-1})q(s_t|o_{\leq t},a_{<t})}{p(h_t|h_{t-1},s_{t-1})p(s_t|h_{t-1},s_{t-1})}] d h_{1:T} d s_{1:T}\\ 
	&= \int_{{h_{1:T},s_{1:T}}}q(h_{1:T},s_{1:T}|o_{1:T},a_{1:T-1}) \\
	&\quad \log [\displaystyle\prod_{t=1}^{T}\frac{q(s_t|o_{\leq t},a_{<t})}{p(s_t|h_{t-1},s_{t-1})}] d h_{1:T} d s_{1:T}.\\  
	\end{aligned}
	\end{equation} 
\end{footnotesize}

Thus:

\begin{footnotesize} 
	\begin{equation} \label{6}
	\begin{aligned}
	&D_{KL}(q(h_{1:T},s_{1:T}|o_{1:T},a_{1:T-1})\parallel p(h_{1:T},s_{1:T}))\\
	&= \int_{{h_{1:T},s_{1:T}}}\displaystyle\prod_{t=1}^{T}q(h_{t}|h_{t-1},s_{t-1})q(s_t|o_{\leq t},a_{<t})\\ 
	&\quad (\displaystyle\sum_{t=1}^{T}\log \frac{q(s_t|o_{\leq t},a_{<t})}{p(s_t|h_{t-1},s_{t-1})}) d h_{1:T} d s_{1:T} \\
	&=\int_{{h_{1:T},s_{1:T}}}\displaystyle\prod_{t=1}^{T}q(h_{t}|h_{t-1},s_{t-1})q(s_t|o_{\leq t},a_{<t}) \\ 
	&\quad (\log \frac{q(s_1|o_1)}{p(s_1)} \\
	&\qquad+ \displaystyle\sum_{t=2}^{T}\log \frac{q(s_t|o_{\leq t},a_{<t})}{p(s_t|h_{t-1},s_{t-1})}) d h_{1:T} d s_{1:T} \\
	&= E_{s_1\backsim q(s_1|0_1)}[\log \frac{q(s_1|o_1)}{p(s_1)}]\\
	&\quad +\int_{{h_{1:T},s_{1:T}}}
	(\displaystyle\prod_{t=1}^{T}q(h_{t}|h_{t-1},s_{t-1})q(s_t|o_{\leq t},a_{<t})) \\
	&\qquad ( \displaystyle\sum_{t=2}^{T}\log \frac{q(s_t|o_{\leq t},a_{<t})}{p(s_t|h_{t-1},s_{t-1})}) d h_{1:T} d s_{1:T} \\
	&=D_{KL}(q(s_1|o_1)\parallel p(s_1))\\
	&\quad +\int_{{h_{1:T},s_{1:T}}}
	(\displaystyle\prod_{t=1}^{T}q(h_{t}|h_{t-1},s_{t-1})q(s_t|o_{\leq t},a_{<t})) \\
	&\qquad ( \log \frac{q(s_2|o_{1:2},a_1)}{p(s_2|h_1,s_1)} \\ &\qquad +\displaystyle\sum_{t=3}^{T}\log \frac{q(s_t|o_{\leq t},a_{<t})}{p(s_t|h_{t-1},s_{t-1})}) d h_{1:T} d s_{1:T} \\
	&= D_{KL}(q(s_1|o_1)\parallel p(s_1))\\ 
	&\quad +\mathbb{E}_{h_1,s_1\backsim q(h_1,s_1|o_1}[D_{KL}(q(s_q|o_{1:2},a_1)\parallel p(s_2|h_1,s_1))] \\
	&\quad +\int_{{h_{1:T},s_{1:T}}}
	(\displaystyle\prod_{t=1}^{T}q(h_{t}|h_{t-1},s_{t-1})q(s_t|o_{\leq t},a_{<t})) \\
	&\qquad ( \displaystyle\sum_{t=3}^{T}\log \frac{q(s_t|o_{\leq t},a_{<t})}{p(s_t|h_{t-1},s_{t-1})}) d h_{1:T} d s_{1:T}.
	\end{aligned}
	\end{equation} 
\end{footnotesize}

Through recursive application of this process to the sum of logarithms indexed by $t$, we obtain:

\begin{footnotesize} 
	\begin{equation} \label{7}
	\begin{aligned}
	&D_{KL}(q(h_{1:T},s_{1:T}|o_{1:T},a_{1:T-1})\parallel p(h_{1:T},s_{1:T}))\\
	&=\displaystyle\sum_{t=1}^{T}\mathbb{E}_{h_{1:t-1},s_{1:t-1}\backsim q(h_{1:t-1},s_{1:t-1}|o_{\leq t-1},a_{<t-1} )}\\
	&\quad [D_{KL}(q(s_t|o_{\leq t},a_{<t})\parallel p(s_t|h_{t-1},s_{t-1}))].
	\end{aligned}
	\end{equation} 
\end{footnotesize}

Finally, we achieve the intended lower bound:

\begin{footnotesize} 
	\begin{equation} \label{8}
	\begin{aligned}
	&\log p(y_{1:T+L},a_{1:T+L}) \\
	& \geq \displaystyle\sum_{t=1}^{T}\mathbb{E}_{h_{1:t},s_{1:t}\backsim q(h_{1:t},s_{1:t}|o\leq t,a<t)}[\log p(y_t|h_t,s_t,\hat{b})+p(a_t|h_t,s_t)] \\   &\quad +\displaystyle\sum_{k=1}^{L}\mathbb{E}_{h_{T},s_{T}\backsim q(h_{T},s_{T}|o\leq T,a<T)}[\log p(y_{T+k}|h_T,s_T,\hat{b})\\
	&\qquad +p(a_{T+k}|h_T,s_T)] \\
	&\quad -\displaystyle\sum_{t=1}^{T}\mathbb{E}_{h_{1:t-1},s_{1:t-1}\backsim q(h_{1:t-1},s_{1:t-1}|o_{\leq t-1},a_{<t-1} )}\\
	&\qquad [D_{KL}(q(s_t|o_{\leq t},a_{<t})\parallel p(s_t|h_{t-1},s_{t-1}))].
	\end{aligned}
	\end{equation} 
\end{footnotesize}

\section{Dataset}  
\label{data}
\begin{figure}[t]
	\centering
	\includegraphics[width=0.45\textwidth]{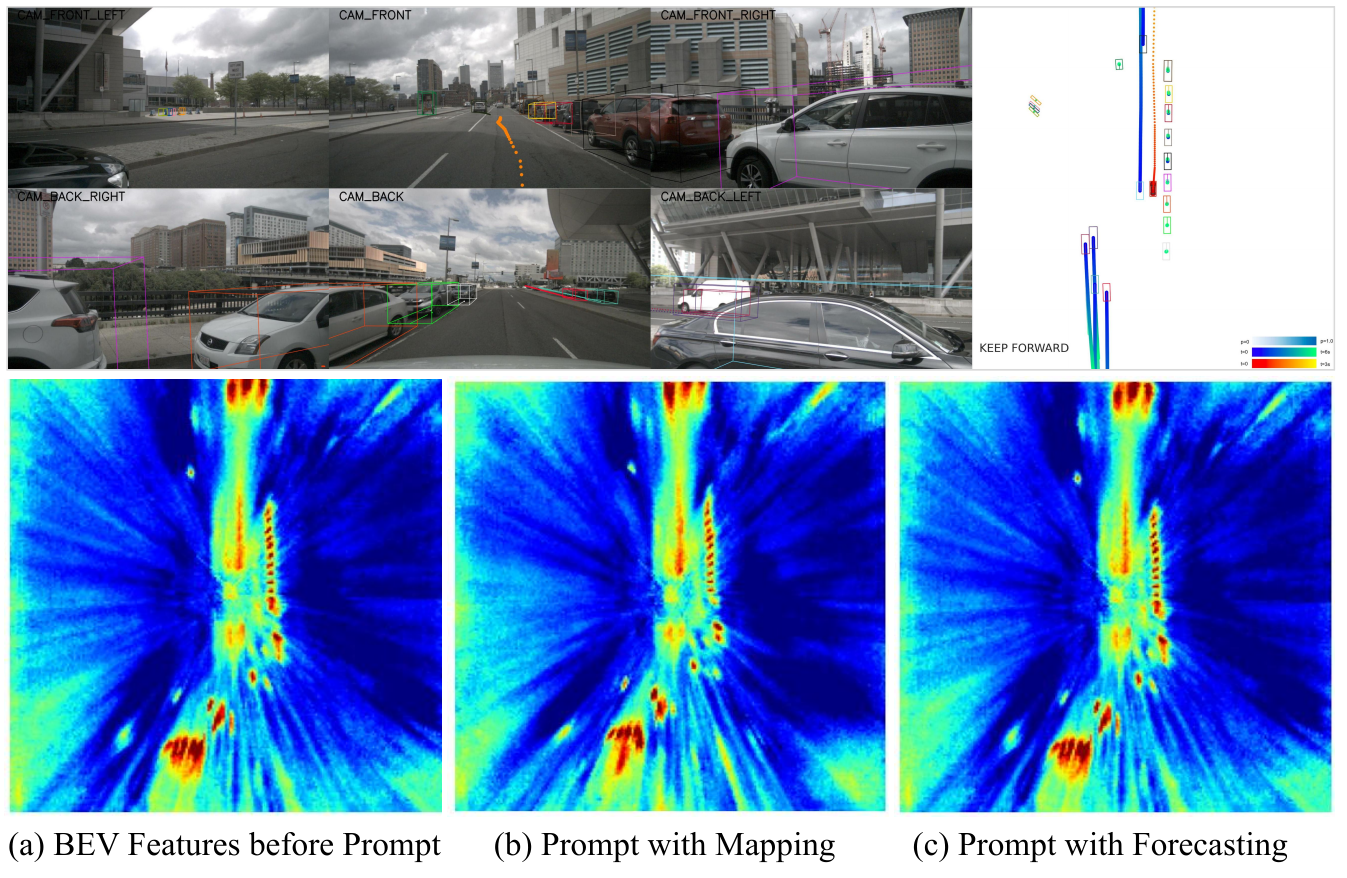} 
	\caption{Visualization of BEV feature maps when prompting with different tasks.}
	\label{fig:task}
\end{figure}

The nuScenes dataset~\cite{nuscenes} is a large-scale autonomous driving dataset that consists of 700, 150, and 150 sequences for training, validation, and testing, respectively. The scenes are recorded in Boston and Singapore, encompassing a diverse array of weather and lighting conditions, as well as various traffic scenarios.

The OpenScene dataset~\cite{openscene} is the largest 3D occupancy dataset, covering a wide span of over 120 hours of occupancy labels collected in various cities, from Boston, Pittsburgh, Las Vegas to Singapore. OpenScene provides a semantic label for each foreground grid and incorporates the motion information of occupancy flow that helps bridge the gap between decision-making and scene representation. we utilize both semantic occupancy labels and occupancy flow for the supervision of 4D pre-training. 

The dense 3D occupancy ground truth is derived by fusing multiple frames of LiDAR point clouds~\cite{occnet,uniscene}. This approach offers a more comprehensive representation of objects, encompassing details about occluded areas, in contrast to single-frame point clouds. In the future, it may become feasible to directly reconstruct 3D occupancy ground truth from autonomous driving videos using techniques such as NeRF~\cite{nerf,nerf2}, 3D Gaussian Splatting~\cite{3dgs}, and MVS~\cite{mvsnet,bi,aa-rmvsnet}.

\section{Task Prompt}
\begin{figure}[t]
	\centering
	\includegraphics[width=0.45\textwidth]{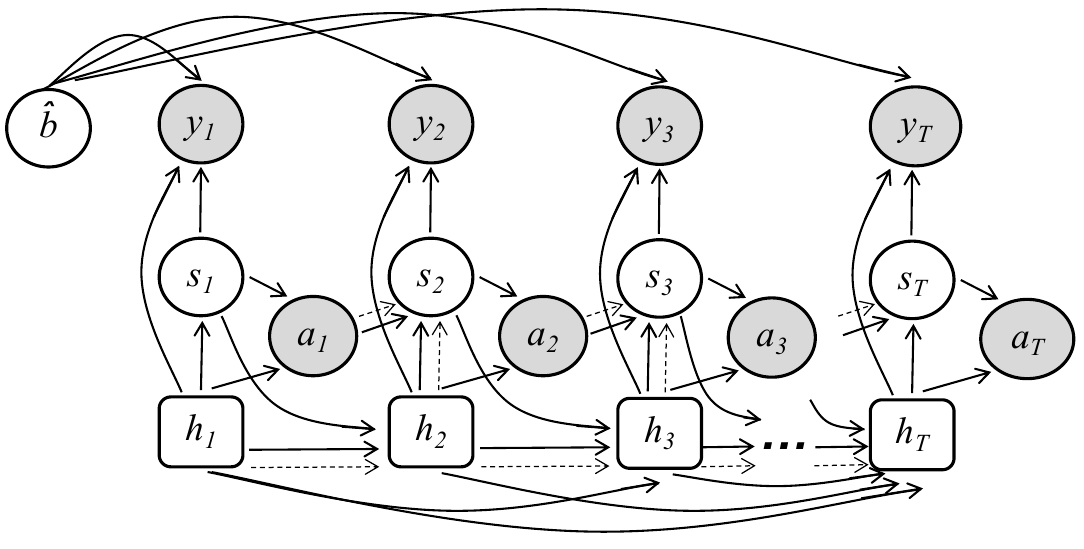} 
	\caption{Graphical model of Memory State-Space Model. Deterministic states are denoted by squares, while stochastic states are represented by circles. The observed states are highlighted in grey for clarity. Solid lines represent the generative model, while dotted lines depict variational inference.}
	\label{fig:graph}
\end{figure}
\begin{figure*}[t]
	\centering
	\includegraphics[width=0.99\textwidth]{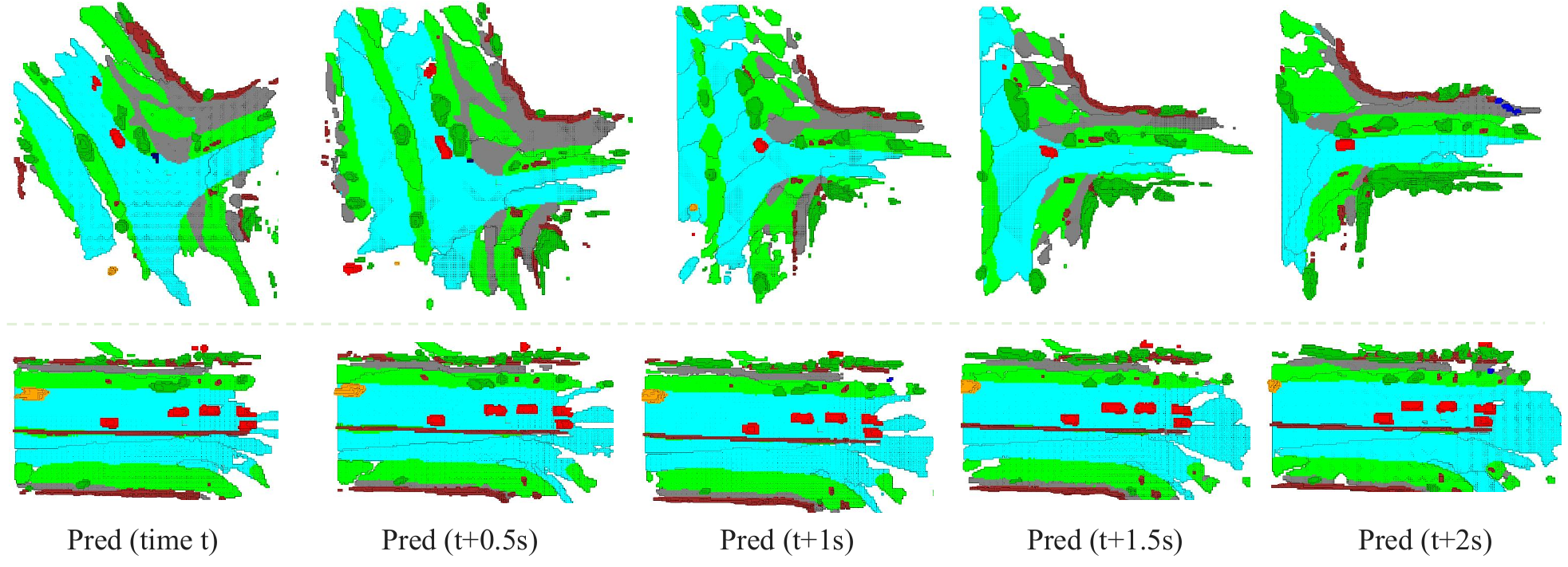} 
	\caption{ Qualitative example of 3D occupancy predictions, for 2 seconds in the future.}
	\label{fig:show}
\end{figure*}
During fine-tuning, we add task prompts to BEV maps before each downstream task's decoder. For 3D object detection, the task prompt is ``The task is for 3D object detection of the current scene." For planning, the task prompt is ``The task involves planning with consideration for both the current and future scenes." The encoder network of task prompts is transferred to downstream tasks, and fine-tuning includes downstream task prompts. This enables different downstream tasks with semantic connections to decouple task-aware features. 
While basic embeddings for specific tasks are optional, large language model captures complex semantic relationships, providing a nuanced representation of task prompts. Additionally, the strong generalization abilities of such models enhance performance across a wide array of tasks when needed.
However, it's worth noting that the current Task Prompt design is relatively simple, and the task number for autonomous driving is limited.

In Fig.~\ref{fig:task}, we present visualizations of BEV feature maps both before and after the integration of various task prompts. 
Notably, as shown in Fig.~\ref{fig:task} (a), the BEV feature map based on 4D pre-training captures abundant information from both the current and future scenes. While, for specific downstream tasks, some information could be redundant or even detrimental. We utilize task prompts to alleviate the effect of redundant information.
In online mapping tasks, the feature map, guided by the task prompt, emphasizes the current spatio-aware information. The targets has more accurate location information in feature map to achieve higher precision .
For motion forecasting task, the feature map, guided by the task prompt, conserves both spatial and temporal information. The targets cover a broader region in feature map to achieve more robust prediction.

\section{Differences between RSSM and MSSM}
In world model-based methods such as Dreamers~\cite{dreamerv1, dreamerv2, dreamerv3} and MILE~\cite{mile}, the RSSM~\cite{latent} is commonly employed to learn latent variables. However, RSSM, relying on RNN networks, may encounter challenges related to long-term information retention. In contrast, our designed Dynamics Memory Bank in MSSM excels in modelling and preserving long-term information. RSSM compresses features into 1D tensor, while MSSM utilizes context BEV features to reconstruct 3D scenes. Besides, MSSM separates dynamic and static information, addressing them independently.

\section{Graphical Model}

In Fig.~\ref{fig:graph}, we illustrate the graphical model of the proposed Memory State-Space Model. The update of the deterministic state $h_t$ is dependent on the historical states in Dynamics Memory Bank, facilitating the transmission of temporal-aware features. Spatial-aware features are preserved through the retention of BEV feature $\hat{b}$. 

\section{Qualitative Results}

Fig.~\ref{fig:show} presents the reconstruction of both the current and future 3D scenes. This visual representation effectively illustrates DriveWorld's capacity for reconstructing the 3D scene and predicting future changes, thus enhancing downstream task performance after 4D pre-training.


\end{document}